\documentclass[lettersize,journal]{IEEEtran}
\usepackage{amsmath,amsfonts}
\usepackage{algorithmic}
\usepackage{algorithm}
\usepackage{array}
\usepackage[caption=false,font=normalsize,labelfont=sf,textfont=sf]{subfig}
\usepackage{textcomp}
\usepackage{stfloats}
\usepackage{url}
\usepackage{verbatim}
\usepackage{graphicx}
\usepackage{cite}
\usepackage{multirow}
\usepackage{color}
\usepackage{tabularx}

\begin{document}

\title{ProIn: Learning to Predict Trajectory Based on Progressive Interactions for Autonomous Driving}

\author{Yinke Dong, Haifeng Yuan, Hongkun Liu, Wei Jing, Fangzhen Li, Hongmin Liu, Bin Fan

\thanks{Yinke Dong, Haifeng Yuan, Hongkun Liu, Hongmin Liu and Bin Fan are with the School of Intelligence Science and Technology, University of Science and Technology Beijing, China (Yinke Dong and Haifeng Yuan contribute equally, Bin Fan is the corresponding author).}
\thanks{Wei Jing is with  Alibaba Group}
\thanks{Fangzhen Li is with alibaba-inc}

}

\maketitle

\begin{abstract}
Accurate motion prediction of pedestrians, cyclists, and other surrounding vehicles~(all called agents) is very important for autonomous driving. \textcolor{black}{Most existing works capture map information through an one-stage interaction with map by vector-based attention, to provide map constraints for social interaction and multi-modal differentiation. However, these methods have to encode all required map rules into the focal agent’s feature, so as to retain all possible intentions' paths while at the meantime to adapt to potential  social interaction.}

\textcolor{black}{In this work, a progressive interaction network is proposed to enable the agent's feature to progressively focus on relevant maps, in order to better learn agents' feature representation capturing the relevant map constraints. The network progressively encode the complex influence of map constraints into the agent’s feature through graph convolutions at the following three stages: after historical trajectory encoder, after social interaction, and after multi-modal differentiation.} \textcolor{black}{In addition, a weight allocation mechanism is proposed for multi-modal training, so that each mode can obtain learning opportunities from a single-mode ground truth.} Experiments have validated the superiority of progressive interactions to the existing one-stage interaction, and demonstrate the effectiveness of each component. Encouraging results were obtained in the challenging benchmarks.

\end{abstract}

\begin{IEEEkeywords}
progressive interaction, multi-modal prediction, trajectory prediction, autonomous driving
\end{IEEEkeywords}

\section{Introduction}
\label{section1}

\IEEEPARstart{T}{rajectory} prediction of traffic agents is an important task for safety autonomous driving as the behavior of traffic agents has a fundamental influence on each other~\cite{trajgan,aandm2a,casas2018intentnet}. However, the various intentions of neighbor agents in scenes are usually unknown. 
In addition, traffic rules encoded in the map have different influence on the agents’ intention, making their interactions even more complicated. 
\textcolor{black}{What is more, trajectory prediction is inherently multi-modal due to its applications, demanding to predict multiple possible trajectories for each agent. However, the scene data only records a single unique ground truth. }
For the above reasons, predicting trajectory of traffic agents~(motion prediction) in the driving scenes is extremely challenging while highly desirable. Many works have been proposed in the past few years to address this problem so as to foster reliable and safety autonomous vehicles.

\begin{figure}[ht]
\centering
\begin{tabular}{cc}
\includegraphics[width=0.48\linewidth]{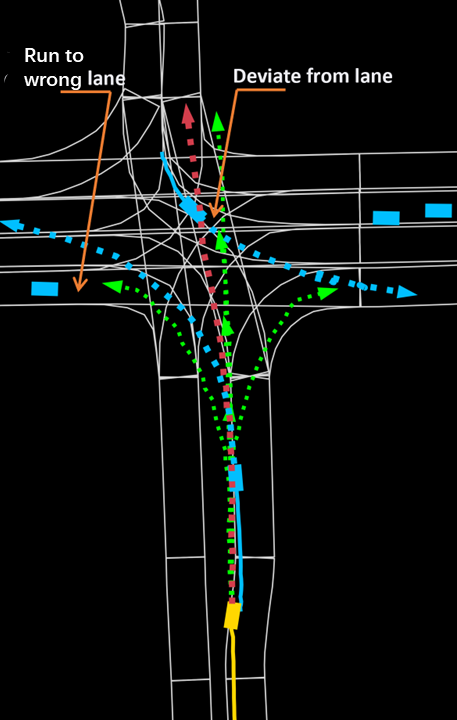} &
\includegraphics[width=0.48\linewidth]{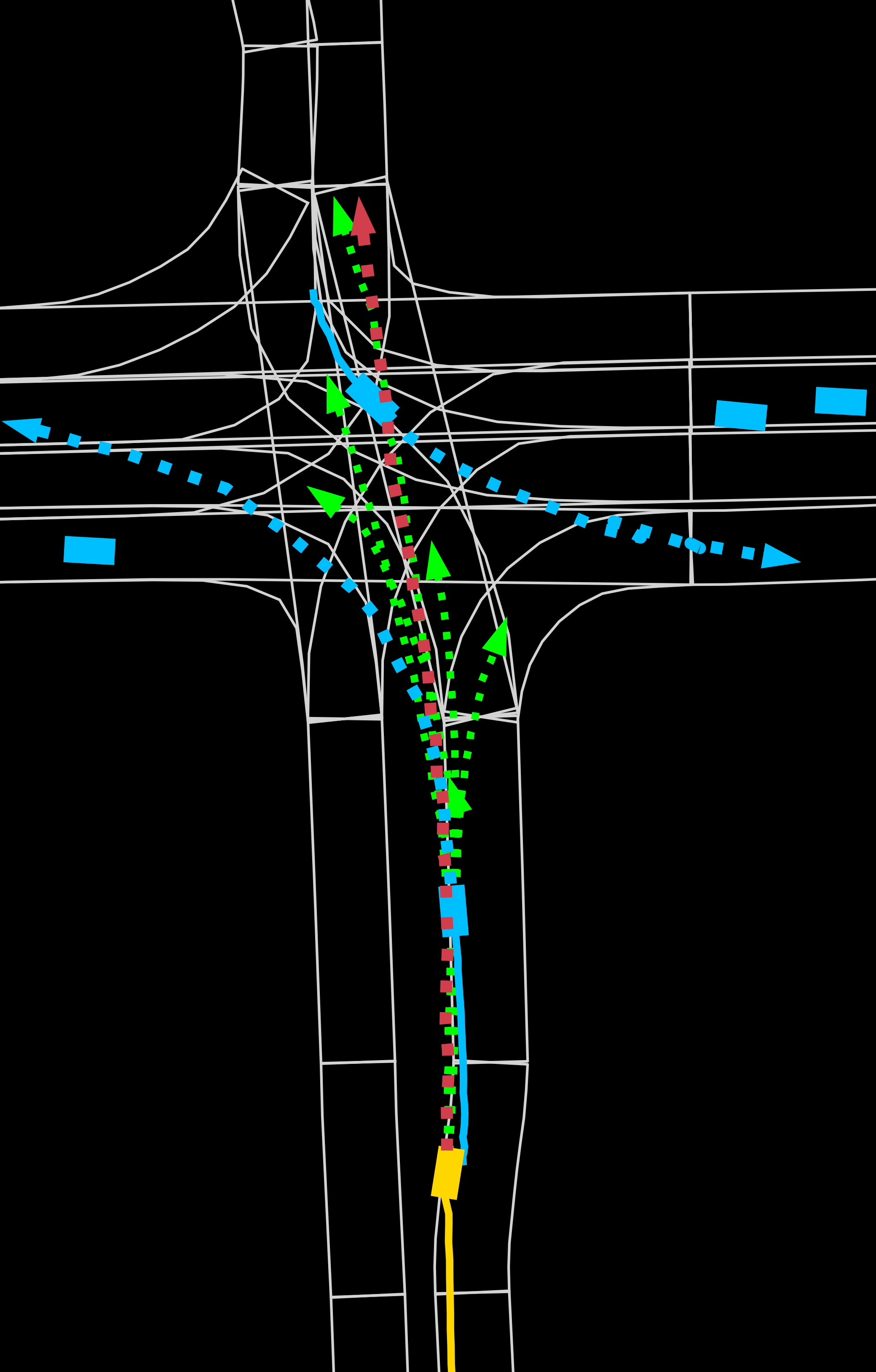} \\
(a)  & (b)  \\
\end{tabular}
\caption{
(a) and (b) are respectively generated by the one-stage and ours in Table \ref{tab:ablation study of M2A}.
(a) displays a failure case where an agent~(yellow) seems to forget the map due to its social interaction with other agents~(blue), resulting in the predicted straight-ahead trajectory~(green) that violates the map rules, and another predicted left turn trajectory~(green) goes to the wrong lane.
(b) presents more reasonable forecasts using our progressive interaction model. 
}
\label{fig:sence_0}
\end{figure}


Existing works for motion prediction can be coarsely divided into two categories: rasterized methods~\cite{casas2018intentnet,hong2019rules,gilles2021home} and vectorized methods~\cite{zhang2022trajectoryDSP, gao2020vectornet, TPA_laneloss, liang2020learning, 2022paga, wang2023ganet, gu2021densetnt, gilles2022gohome, HeteroGCN, liu2021multimodal}. 
\textcolor{black}{The rasterized methods~convert traffic scenes into bird-eye view (BEV) images and then exploit convolutional neural networks (CNNs) for learning. These methods usually suffer from information loss and are inefficient in capturing topology structure in the map with 2D convolutions. }
\textcolor{black}{The vectorized methods have garnered increasing attention from researchers in recent years.
They encode maps, agents, and other scene elements as vector features, and then rely on permutation-invariant set operators (e.g., point cloud convolutions~\cite{ye2021tpcn, ye2023dcms}, graph convolutions~\cite{liang2020learning, 2022paga}, and Transformers~\cite{vaswani2017attention}) to capture scene context for trajectory prediction. 
These methods utilize designed Graph Convolutional Networks~(GCNs) to model map graph~\cite{liang2020learning, 2022paga}, incorporate explicit map rule constraints into feature learning~\cite{TPA_laneloss,Xiaoyu,laneheadloss, nayakanti2022wayformer}, or derive map-related anchors\cite{zhang2022trajectoryDSP, gu2021densetnt} for motion prediction. All in all, these methods capture scene context by encoding map rules into the agent’s feature all through an one-stage interaction between the agent and maps. For the reason that social interaction and mode differentiation also require map rules, such one-stage interaction occurs before the social interaction module and mode differentiation module in most methods. 
}

\textcolor{black}{However, this way of interaction faces a challenging problem of feature representation learning for the focal agent, including the following two aspects: (i) on the one hand, agent feature needs to encode and capture a wide range of map areas so as to adapt all possible social influence of other agents as well as the focal agent’s multiple intentions. (ii) on the other hand, in the subsequent social interaction module and multi-modal differentiation module, the agent feature needs to further consider how to retain map constraints. Therefore, the network needs to ensure that the informative map constraints are not lost during previous steps. 
}

\textcolor{black}{To alleviate the above problems, we propose a novel \textbf{Pro}gressive \textbf{In}teraction network~(called ProIn),
which progressively encode the complex influence of map constraints into the agent’s feature through graph convolutions at three stages, i.e., 
after historical trajectory encoder, after social interaction, and after multi-modal differentiation.
This progressive interaction network has the following advantages. Firstly, good feature representation capturing the intrinsic map constraints is easier to learn, as agents only need to focus on the regions relevant to the current stage, resulting in more concentrated map constrains. Furthermore, during social interaction and multi-modal differentiation, there is no need for the agent feature to consider map information as it has already been properly encoded. These supplemented map constraints can help the focal agent to choose areas where it originally has a low probability of passing through, if not influenced by neighboring agents. Finally, they are also beneficial for the focal agent to focus on the chosen road related map after it determines the explicit intent. As illustrated in Figure~\ref{fig:sence_0}, our model outputs more reasonable forecasts compared to the one-stage methods.
}

\textcolor{black}{In addition, as a widely used training strategy widely used in multi-modal prediction, winner-take-all (WTA) strategy only uses the closest mode to calculate the loss regarding the ground truth~\cite{liang2020learning, zhang2022trajectoryDSP, 2022paga, wang2023ganet, HeteroGCN, huang2022multi, TPA_laneloss}. }
\textcolor{black}{This may lead to mode collapse and sample inefficiency\cite{ye2023dcms,makansi2019overcoming, gilles2022gohome}. We empirically compare the predictive ability of each modal branch and observe a significant disparity in terms of proportion and accuracy (please refer to Table III).} To conquer this problem, we design an allocation loss, enabling to change the learning strength of each modal branch according to the prediction of the sample. The better the prediction is, the greater the learning weight will be assigned to the mode. In this way, for each sample, all modes can be trained with different weights.

To sum up, our contributions are two-folds:

$\bullet$ A progressive interaction network is proposed to enable agents gradually acquiring map information at different levels through multi-stage interactions.

$\bullet$ A weight allocation mechanism is proposed for the multi-modal prediction training in autonomous driving so that a single-mode ground truth can be used to train for multi-modal output trajectories.


\section{Related work}
\textcolor{black}{\textbf{Scene Context Learning in Motion Prediction:}
In vectorized methods, the agent and traffic elements like lanes, signal lights, etc. are represented using various vectors by 1D CNN~\cite{liang2020learning}, LSTM~\cite{huang2022multi} or Transformers~\cite{zhou2022hivt}. These irregular vector representations require permutation-invariant set operators to aggregate context information. }

TPCN\cite{ye2021tpcn} and DCSM\cite{ye2023dcms} inherits from point cloud learning methods, utilizing point-wise and voxel-wise dual-representation to model spatial relationships and temporal dependencies. 
SceneTransformer\cite{ngiam2022scene} utilizes Transformers along the temporal and spatial axes to model context. Wayformer\cite{nayakanti2022wayformer} connects different inputs into the same structure to achieve unified multi-axis attention in the Transformers. Macformer\cite{MacFormer} explicitly integrates map constraints by the designed coupled map module and reference extractor. 
ProphNet~\cite{Wang_2023_CVPR} introduces a uniform and succinct transformer under the agent-centric data representation to achieve higher prediction accuracy.
QCNet~\cite{Zhou_2023_CVPR} achieves streaming scene encoding with a novel query-centric data representation.

Some recent works\cite{zhang2022trajectoryDSP, gao2020vectornet, TPA_laneloss, liang2020learning, 2022paga, wang2023ganet, gu2021densetnt, gilles2022gohome, HeteroGCN } utilize Graph Convolutional Networks (GCNs) to model the irregular topology of roads and relationships between agents and map elements. 
VectorNet~\cite{gao2020vectornet} models interactions among lanes and agent trajectories with GCNs. LaneGCN~\cite{liang2020learning} builds a lane graph on the basis of lane nodes and uses multiscale graph convolution networks to learn node features. PAGA\cite{2022paga} further extends the second-order edge relations in LaneGCN. DSP\cite{zhang2022trajectoryDSP} and GANet\cite{wang2023ganet} rely on generating anchors of scene map to enhance the agent features. HeteroGCN\cite{HeteroGCN} employs a dynamic heterogeneous graph convolutional recurrent network to aggregate dynamically changing interactions over time and capture their evolution. 
Furthermore, some methods\cite{TPA_laneloss,Xiaoyu,laneheadloss} introduce various explicit constraints to help agent learn map rules.

In contrast with these methods, we propose a novel progressive interaction module to capture the scene context information, enabling agents to capture map constraints suitable for social interaction and multi-modal intentions.

\textcolor{black}{\textbf{Multi-modal Prediction: }
Motion forecasting is inherently modeled as a multi-modal prediction task due to the agents’ highly uncertain intentions and the safety requirement of autonomous vehicles. 
Goal-based\cite{gu2021densetnt, zhang2022trajectoryDSP, gilles2022gohome} methods select a subset of goals from a large pool or a heatmap of agent goals to mitigate the risk of modal averaging. Many works\cite{liang2020learning, zhang2022trajectoryDSP, 2022paga, wang2023ganet, HeteroGCN, huang2022multi, TPA_laneloss} directly train multiple regression heads with WTA training strategy, albeit the risk of mode collapse and sample inefficiency\cite{makansi2019overcoming, gilles2022gohome}. 
Another type of works\cite{makansi2019overcoming, zhou2022hivt, MacFormer, nayakanti2022wayformer, gilles2021thomas} estimates a parametric mixture distribution of trajectories by optimizing the negative log-likelihood loss for sampled trajectories, like GMM-based methods~\cite{makansi2019overcoming,MacFormer}. Due to the intractability of directly fitting mixture model likelihoods\cite{makansi2019overcoming}, these methods still update only one selected component for each training sample. Our proposed two-stage allocation loss allows to update all modes in different extents for each training sample, effectively improving the weaker branches in the WTA training strategy. }

\section{The proposed method}

\begin{figure*}[htp]

    \centering
    \includegraphics[width=16.5cm]{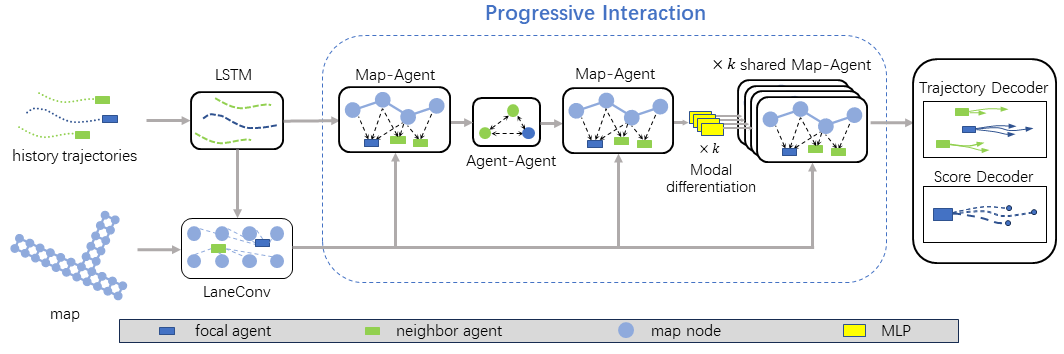}
    \caption{The pipeline of our method. It first extracts features of agents and map independently, then uses GCNs to implement a series of interactions between agents and map, and finally uses six branches to generate multi-modal trajectories.}
    \label{fig:galaxy}
\end{figure*}

\subsection{Problem Formulation}
\label{section3.1}
For a given agent, {denote its} history trajectory including $T$ time steps as $ \mathbf{p}(t), t =1, \cdots, T $, where  $ \mathbf{p}(t) \in \mathbb{R}^{2} $ is the position of the agent at time $t$. As in previous work~\cite{gao2020vectornet}, this history trajectory of the agent is preprocessed into displacements. A third dimension with 1 or 0 is appended to the displacement to indicate whether this position is valid or not. As a result, the trajectory of the $i$-th agent is finally represented by $X_i = [ \mathbf{x}_i; \mathbf{b}_i ] \in \mathbb{R}^{3 \times T}$, where $ \mathbf{x}_i = [\mathbf{0}, \mathbf{p}_i(2) - \mathbf{p}_i(1), \cdots,  \mathbf{p}_i(T) - \mathbf{p}_i(T-1) ] \in \mathbb{R}^{2 \times T}$ and $\mathbf{b}_i \in \{1,0\}^{1 \times T}$, as input to our method. Note that if any item in $\mathbf{p}(t) - \mathbf{p}(t-1)$ is missing, its value is set as $\mathbf{0}$ in $\mathbf{x}$. The subscript $i$ denotes the agent. Similarly, the map is denoted by a graph of lane segments $\mathbf{L}_i = [ \Delta x_i, \Delta y_i, Cx_i, Cy_i ], i=1,2,\cdots,n$, where $(\Delta x, \Delta y)$ are the displacements computed by subtracting the start point of the lane segment from its ending point and $(Cx, Cy)$ are the position of its central point. There could be additional binary vector appended to the lane segment to describe the corresponding traffic rules of this segment such as the turn type, intersection or not, etc. In the map graph, each lane segment is a vertex, and one directed edge is defined to connect lane segments if they are reachable according to the traffic rules. 
The task of motion prediction is to predict the future positions of a given agent~(called focal agent later in this paper) for the next $F$ time steps based on the map and trajectories of all agents in the scene. 
There are $K$ outputs, known as multi-modal outputs, denoted as
$ Y_{i}^{k} = \{ \mathbf{y}_{i}^{k}, s_{i}^{k} \}, k = 1,\cdots,K $, where $\mathbf{y}_{i} ^{k}= [ \mathbf{p}_{i}^k(T+1), \cdots, \mathbf{p}_{i}^k(T+F) ]$ is the predicted future trajectory, and the superscript $k$ represents the $k$-th output and $s^k$ is the corresponding confidence score. Note that all the positions of agents and lane segments in a scene are transformed to a BEV coordinate system~\textcolor{black}{\cite{liang2020learning}}, which is defined by the position and speed direction of the focal agent.

\subsection{Overall Framework}

Figure \ref{fig:galaxy} shows an overview of the proposed framework. Firstly, the agent history and map represented by vectors described in Section \ref{section3.1} go through two encoders respectively to extract their features~(Section \ref{section3.3}). Secondly, the agent features are enhanced by the proposed progressive interaction network~(Section \ref{section3.4}). 
There are four modules in the progressive interaction network, three of which use graph convolutions to model the interactions of map and agents~(i.e., the Map-Agent in Figure~\ref{fig:galaxy}) and the remaining one models the interactions among agents~(the Agent-Agent in Figure~\ref{fig:galaxy}). The enhanced features are differentiated into $K$ branches with Multilayer Perceptron (MLP) to predict multi-modal outputs. Finally, each of the $K$ feature branches is taken as input of $K$ MLPs to output future trajectories, which are concatenated with the agent feature to obtain the corresponding scores. 
The learning of this network is supervised by losses defined on the quality of outputs~(Section \ref{section3.5}).

\subsection{Encoder}
\label{section3.3}
{\bf Agent Encoder: } To extract the spatial-temporal feature embeddings of agent history, we use a single-layer LSTM to obtain features with $d_a$ channels of $T$ moments. Features of all time steps are combined through a fully connected layer to obtain a feature with dimension $d_a$ as the output of encoder.

{\bf Map Encoder: } The map is represented by a graph of lane segments, whose feature embeddings are learned through a graph convolution network~(i.e., the LaneConv in~\cite{liang2020learning}). For each lane segment, its central position, shape defined by the displacement, additional binary vector~(if any) encoding the traffic rules go through three MLPs to extract the corresponding features. These features along with the agent features in a neighborhood of the lane segments are merged as the features of the vertex in the lane graph. The final lane segment features are extracted through a graph neural network with three LaneConv layers. More concretely, we use smaller forward and backward expansions~(LaneConv(1, 2, 4)) instead of LaneConv(1, 2, 4, 8, 16, 32) in~\cite{liang2020learning} to reduce the computation cost for extracting map features.

\subsection{Progressive Interaction Net}
\label{section3.4}
\subsubsection{Graph Convolution for Interaction}

{\bf Map-Agent Interaction} To incorporate map information in the agent feature, we jointly consider the features of agent and map node and their spatial distance in a graph convolution: 
\begin{equation}
  \mathbf{g}_{ij} = \varphi_{2}  (concat[  \mathbf{a}_{i} W_{1}, \mathbf{ m}_{j} ,\varphi_{1}( \Delta d_{ij} )])
  \label{eq:3}
\end{equation}
\vspace{-8pt}
\begin{equation}
\mathbf{a}_{i}'  =  \mathbf{a} _{i}W_{2} +\sum_{ j\in \mathcal{M} _{i}  } softmax ( \varphi  _{3}(\mathbf{g} _{ij})  ) \odot   \mathbf{g} _{ij} W_{3} 
  \label{eq:4}
\end{equation}

where $ \mathbf{m}_{j}$ is the $j$-th map node feature, $\mathbf{a}_{i}$ is the input agent feature and $\Delta d_{ij}$ denotes the spatial distance between the agent and map node. $W_{1}$, $W_{2}$, $W_{3}$ are trainable weights for linear feature embedding and $\varphi_{1}$, $\varphi_{2}$, $\varphi_{3}$ are two-layer MLPs with 128 hidden units and 128 output nodes. $\mathcal{M} _{i}$ denotes the neighborhood of the $i$-th agent. By the above graph convolution, $\varphi_{1}$ extracts the feature of relative position between agent and map, and $\varphi_{2}$ generates the relation vector used for attention learning. For simplicity, this graph convolution update process can be expressed as $\mathbf{a}' _i= \mathcal{G}(\mathbf{a} _i,\mathbf{m} _j, \Delta d_{ij}) , j \in \mathcal{M} _i$.

\begin{figure}[h]
    \centering
    \includegraphics[width=7cm]{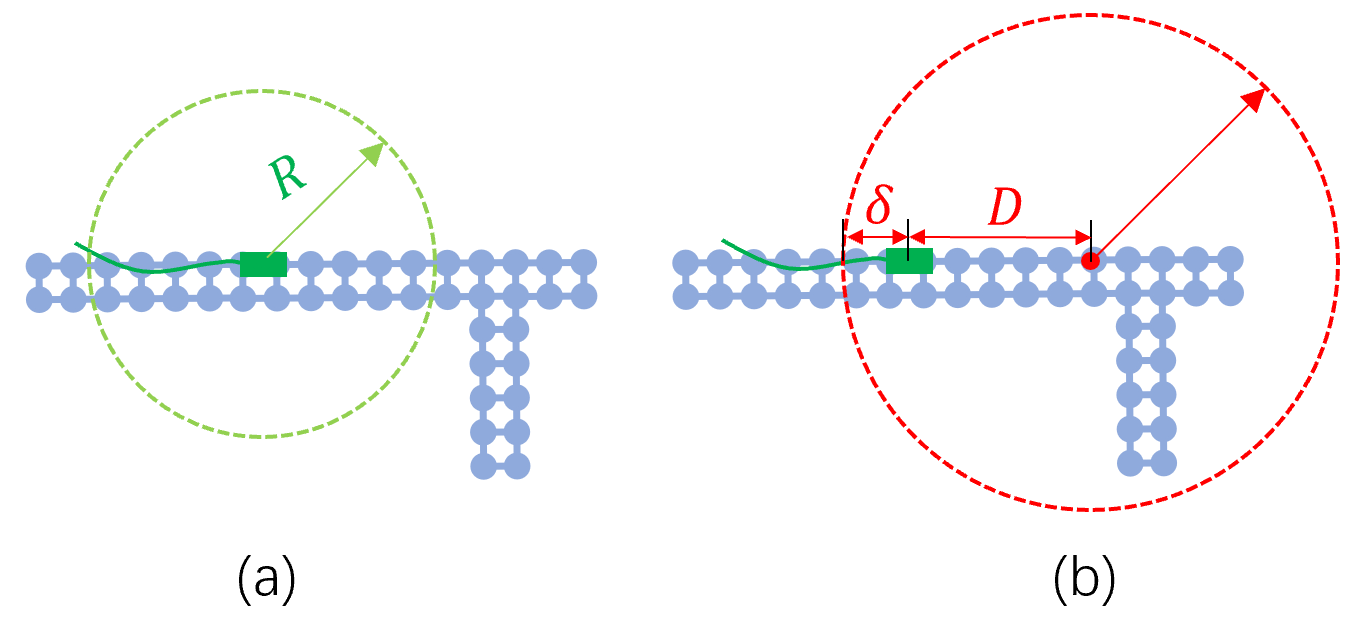}
    \caption{Illustration of neighbors in Map-Agent interaction with (a) fixed range and (b) the proposed dynamic range, where $\mathbf{p}(T)+D$ is set as the center of the circle with radius of $|D|+ \delta$.}
    \label{fig:nodes2a}
\end{figure}

A key issue for the above graph convolution based Map-Agent interaction is to select appropriate neighboring map nodes for a given agent. Instead of statically delimiting a fixed range (see Figure~\ref{fig:nodes2a}(a)) as neighbors, we adopt a dynamic range selection strategy. \textcolor{black}{The idea is that the agent needs to pay more attention to the area along its moving direction, and the interaction range should be adapted to the speed and prediction time duration.} Therefore, given an agent, we take its position $\mathbf{p}(T)+D$ as the center of neighbor, and the range is set by a circle with $|D|+ \delta$ as the radius, as demonstrated in Figure~\ref{fig:nodes2a}(b). $D$ is set as the dynamic vector that the current speed moves forward by 25-time steps 
in our experiments. This dynamic strategy is not only more effective but also contributes to faster inference speed, since there are many low-speed agents whose neighborhoods can be smaller with less nodes in Map-Agent to compute.

{\bf Agent-Agent Interaction}
Similar to the Map-Agent interaction, the graph convolution to learn the relationship between the $i$-th agent and it's neighbors can be described as $  \mathbf{a}' _i= \mathcal{G}(\mathbf{a} _i,\mathbf{a} _j, \Delta d_{ij}) , j \in \mathcal{A} _i$, where $ \Delta d_{ij}$ donates the spatial distance of two agents and $\mathcal{A} _{i}$ denotes the set of neighboring agents of the $i$-th agent. Due to the limited number of agents in the scene, we use a fixed radius to define $\mathcal{A} _{i}$ with a radius 100m to acquire enough neighboring agents.


\subsubsection{Progressive Agent Feature Learning}
\textcolor{black}{Agents’ behaviors are heavily influenced by their social interactions with surrounding agents and traffic rules encoded in the map features. Due to unknown intentions and uncertainties in future, the reasonable actions that agents can take may vary significantly. 
\textcolor{black}{Especially, in challenging driving scenarios, it is difficult to obtain sufficient map information for perception social behavior and multiple intentions through an one-stage interaction with map. }
A progressive interaction mechanism is thus proposed to better encode map information adapted to social interaction and multi-modal prediction into the agent feature, improving the encoding of contextual information in the scene. The mechanism includes four modules.}

\textcolor{black}{Firstly, after historical trajectory encoder, we use a {\bf Map-Agent} interaction module to enable agents to plan their driving intentions considering both the map information and historical trajectories. Secondly, as the map is only one of many factors when agents plan for their future actions, predicting their future motions needs to completely consider the social relations of other agents in the scene. An {\bf Agent-Agent} interaction module is used to capture this social relationship so as to consider the possible future trajectories of neighboring agents in the map. 
Thirdly, to model the fact that agents may change their initial routing after taking full account of social relations, another {\bf Map-Agent} interaction module is added after the {\bf Agent-Agent} interaction module. This module is important as it enables agents to choose areas with originally low probabilities of passing through. 
Finally, we use k independent MLPs with residuals to achieve multi-modal differentiation. After differentiation, each branch models one possible clear motion intent based on the previous extracted features and is responsible for generating the final predicted trajectory. A third {\bf Map-Agent} interaction module help the agent specifically focus on the map related to the chosen route, to accurately predict a trajectory. }
The whole procedure of progressive interactions described above is illustrated in Figure~\ref{fig:galaxy}. Through this procedure, the map information is actively and adequately fused into agent features, which we believe is reasonable based on our intuitions about decision making in driving scenarios and can not be easily learned in a data-driven way through a plain graph neural network. The superior performance to other graph based methods and ablations demonstrated in the experiments also validate the effectiveness of this progressive interaction mechanism.





\subsection{Training}
\label{section3.5}

The learning objectives of our method contain three kinds of loss terms $ \mathcal L = \mathcal L _{cls} + \mathcal L _{reg}$, where $\mathcal L _{cls}$ is for training the scoring head, and $\mathcal L _{reg}$ is the loss for training the trajectory generation head. 
For the scoring head, we use the max-margin loss. In addition, since we want to minimize Brier-minFDE~\cite{chang2019argoverse}, we add $ (1-s^{\hat{k}})^{2} $ to the scoring loss function, where $s^{\hat{k}}$ is the score corresponding to the best trajectory. Therefore, the $\mathcal L _{cls}$ can be written as:
\begin{equation}
\mathcal L _{cls}= \frac{1}{NK} \sum_{i} \sum_{k\ne \hat{k} } max \left (   0,s_{i}^{k} + \epsilon - s_{i}^{\hat{k}} \right ) + \eta \cdot (1-s_{i}^{\hat{k}})^{2} 
  \label{eq:cls}
\end{equation}
\vspace{-2pt}
where $\eta$ is a hyper parameter set as 2.0, $ k $ represents the $k^{th}$ mode, and $\hat{k}$ represents the mode with the minimal endpoint error, $N$ is the total number of agents.

For the multi-model trajectory regression head, we only have a single-mode ground truth trajectory. The widely used training method is the winner-take-all strategy, which only computes gradient based on one prediction branch. This way has the risk of insufficient training. \textcolor{black}{As shown in Table \ref{tab:allocation of 6 modalities}, the modal branch 4, 5, and 6 only use $3\% \sim 15\%$ of the data for training, and a severe lower accuracy is observed for these branches. 
}
 To conquer this, we design a two-stage training strategy to train every branch with a single-mode ground truth.

In the first training stage, the $smoothL1$ loss, denoted as $\phi(\cdot)$, is used to warm up the network, so the regression loss $\mathcal {L} _{reg}$ at this stage is, 

\begin{equation}
\mathcal L _{reg} = reg_{L_1}\left ( \mathbf{y} \right ) = \frac{1}{NF} \sum_{i}\sum_{t} \phi \left ( \mathbf{p}_{i}^{\hat{k}}(t) - \mathbf{p}_{i}^{*}(t)  \right ) 
  \label{eq:L_1}
\end{equation}
\vspace{-1pt}
where $\mathbf{p}$ is the predicted trajectory and $ \mathbf{p}^{*} $ is the corresponding ground truth, $t= T+1, \cdots, T+F$ is the future time step, for simplicity, we denote the final time step $(T+F)$ as $\mathcal{T}$.

In the second stage, we use an allocation loss to update all prediction branches, including those whose output are not the closest to the ground truth trajectory. 
\begin{equation}
reg_{a}\left ( \mathbf{y} \right ) = \frac{1}{NFK} \sum_{i}\sum_{t}\sum_{k} w^{k} \cdot \phi \left ( \mathbf{p}_{i}^{k}(t) - \mathbf{p}_{i}^{*}(t)  \right ) 
  \label{eq:reg2}
\end{equation}
\vspace{-1pt}
In Eq.~(\ref{eq:reg2}), a weight $w^{k}$ is assigned to each mode based on their final displacement errors, so as to allocate ground truth to each mode. The smaller endpoint error of the mode is, the greater learning weight will it be. The allocation weight $w^{k}$ is determined by the following equation,
\begin{equation}
w^{k}=softmax(\mathcal{F}( \mathbf{p}^{k}(\mathcal{T} )-{\mathbf{p}^{*}}(\mathcal{T} )))
  \label{eq:7}
\end{equation}
where $\mathcal{F}(\mathbf{\Delta p} ) = \frac{\zeta }{\varsigma + \left \| \mathbf{\Delta p}  \right \|_{2}^{2} }$ is not back-propagated, $\zeta$ and $\varsigma$ are set as 8.0 and 4.0 in our experiment. 

Due to the importance of endpoints in the prediction task, we add an endpoint loss, which is denoted as Eq.~(\ref{eq:reg3}).
 \begin{equation} 
reg_{e}\left ( \mathbf{y} \right ) = \frac{1}{N} \sum_{i}\Vert \mathbf{p}_{i}^{\hat{k}}(\mathcal{T} )-\mathbf{p}_{i}^{*}(\mathcal{T} )\Vert_{2}^{2} 
  \label{eq:reg3}
\end{equation}

So, the second stage regression loss is written as,
\begin{equation}
\mathcal L _{reg}  = reg_{L_1}(\mathbf{y})   + \alpha * reg_{a}\left ( \mathbf{y} \right )   + \beta * reg_{e}\left ( \mathbf{y} \right )
  \label{eq:reg_s2}
\end{equation}
where $\alpha$, $\beta$ are set as 1 and 0.2 in our experiments.

\section{Experiments}
\label{sec:formatting}
\subsection{Experimental setup}

{\bf Dataset: }We evaluate our method on the large-scale Argoverse 1 motion forecasting dataset~(AV1)\cite{chang2019argoverse} and Argoverse 2 motion forecasting dataset~(AV2)\cite{Argoverse2}. 
AV1 contains over 300K real-world driving scenes, providing splits of training, validation, and test sets. 
Each sequence includes map information and agent trajectories, each of which has a duration of 5 seconds and is sampled into 50-time steps. 
According to the given 2 seconds history for the test set, it requires to predict the motion trajectory in the next 3 seconds.
AV2 contains 250K scenes sampled in six cities and provides more information on agents and map. The agent history includes speed, location, and the agent type, which include most static and dynamic agents. The map gives the driveable area, crosswalk, and lane information which includes the center line, lane type, left and right boundaries, boundary line type, as well as lane topological relationship. According to the given 5 seconds history for the test set, it needs to predict the trajectory in the next 6 seconds.


{\bf Metrics: }\textcolor{black}{According to the setting of Argoverse and Argoverse2 Motion Forecasting leaderboards\cite{argo.org, argo2.org}, we predict K=6 trajectories for each focal agent, and compute the folowing metrics: Average Displacement Error~(minADE), minimum Final Displacement Error~(minFDE), Miss Rate~(MR) and Brier-minFDE. ADE is the averaged Euclidean distance between the prediction and the ground truth over all time steps, and minADE refers to the minimum ADE of $K$ predictions. }
\textcolor{black}{minFDE represents the minimal Euclidean distance between the predicted position at final time step and the ground truth among $K$ predictions. 
Brier-minFDE is the sum of minFDE and the prediction confidence in multi-mode prediction tasks. MR is defined as the ratio of scenarios where the minFDE is beyond 2 meters. As minFDE is not averaged over other time steps, it reflects how the \emph{consistent} high quality trajectory one method can achieve across all the time steps. minFDE and Brier-minFDE are also considered as the two most important metrics for comparing different methods based on Argoverse and Argoverse2~\cite{chang2019argoverse,Argoverse2}. }

\textcolor{black}{In addition, to study the influence of the proposed allocation loss on each prediction branch, we propose {\bf Hit Rate} as an additional metric. Hit Rate indicates the proportion of scenes, where the predicted trajectory from a branch is the closest one to the ground truth among $K$ branches.}


{\bf Implementations: }We apply some standard data augmentation during the training stage, including random flipping along the x-axis with a probability of 0.3 and random mask the history trajectories at the first 30\% with a probability of 0.3. 
As explained in Section \ref{section3.5}, we use a two-stage training strategy. For the first stage of training, the learning rate is set to 0.001 and 0.0001 to train the network for 32 epochs and 2 epochs respectively, using the regression loss defined in Eq.(\ref{eq:L_1}). For the second stage of training, the network is trained with 46 epochs and 10 epochs with a learning rate of 0.001 and 0.0001, respectively, using the regression loss of Eq.(\ref{eq:reg_s2}). 
For both stages, the scoring loss is described in Eq.(\ref{eq:cls}), the batch size is set to 32, and the optimizer uses the Adam~\cite{kingma2014adam}.  

To evaluate on AV2, we adjust the model to adapt to the input data of AV2. First, we modify the time length of the Encoder and Decoder. 
Then, additional type information in the dataset is encoded into the input of the Encoders by adding learnable embedding layers. 
Since agents need to know the map with a larger distance, the receptive field of LaneConv is doubled to LaneConv(1,2,4,8), and the number of layers in Map-Agent interaction modules are doubled accordingly.

\subsection{Ablation study}

{\bf Component study. }
To analyze contributions of each component, we conducted various ablations and reported the results in Table \ref{tab:ablation study of M2A}. For neatness, we denote the Agent-Agent interaction component as A2A and mark the three Map-Agent interaction at different stages as M2A$_{e}$ (after the encoder), M2A$_{s}$ (after the social interaction), and M2A$_{m}$ (after the multi-modal differentiation). 

\begin{table}[ht]
  \caption{Ablation study of each component on AV1 test set}
  \centering
    \scalebox{0.85}{
  \begin{tabular}{c|c|cccc|cccc}
    \hline
\makebox[0.02\textwidth][c]{No.} & Method             & \makebox[0.035\textwidth][c]{M2A$_{e}$}&\makebox[0.035\textwidth][c]{A2A}      & \makebox[0.035\textwidth][c]{M2A$_{s}$}&\makebox[0.035\textwidth][c]{M2A$_{m}$} &minFDE & \makebox[0.035\textwidth][c]{b-minFDE} & MR\\
    \hline
1   & \textcolor{black}{one-stage}        & $\surd$  & $\surd$  &          &          & 1.227 & 1.820 & 13.1 \\  

2   & \textcolor{black}{w/o M2A$_{m}$}    & $\surd$  & $\surd$  & $\surd$  &          & 1.188 & 1.774 & 12.2 \\  

3   & \textcolor{black}{w/o M2A$_{s}$}    & $\surd$  & $\surd$  &          & $\surd$  & 1.178 & 1.770 & 12.0 \\  

4   & \textcolor{black}{w/o A2A}          & $\surd$  &          & $\surd$  & $\surd$  & 1.227 & 1.818 & 13.3 \\  

5   &\textcolor{black}{w/o M2A$_{e}$}     &          & $\surd$  & $\surd$  & $\surd$  & 1.186 & 1.767 & 12.5 \\  
    
6   & \textcolor{black}{ours}             & $\surd$  & $\surd$  & $\surd$  & $\surd$  & {\bf 1.163} & {\bf 1.750} & {\bf 11.6} \\  



    \hline
  \end{tabular}
  }

  \label{tab:ablation study of M2A}
\end{table}

\textcolor{black}{In Table \ref{tab:ablation study of M2A}, one-stage has basic interaction of map and agent. By comparing the first three rows in Table~\ref{tab:ablation study of M2A}, it is apparent that each stage of incorporating map information into the agent features contributes to the performance improvement to a certain extent. By removing any module in our full model, it leads to degraded accuracy as demonstrated by the results of the second to fifth rows. These ablations clearly validate that each of the proposed component makes a non-negligible contribution to the final performance. }
\begin{figure*}[htp]

    \centering
    \includegraphics[width=16cm]{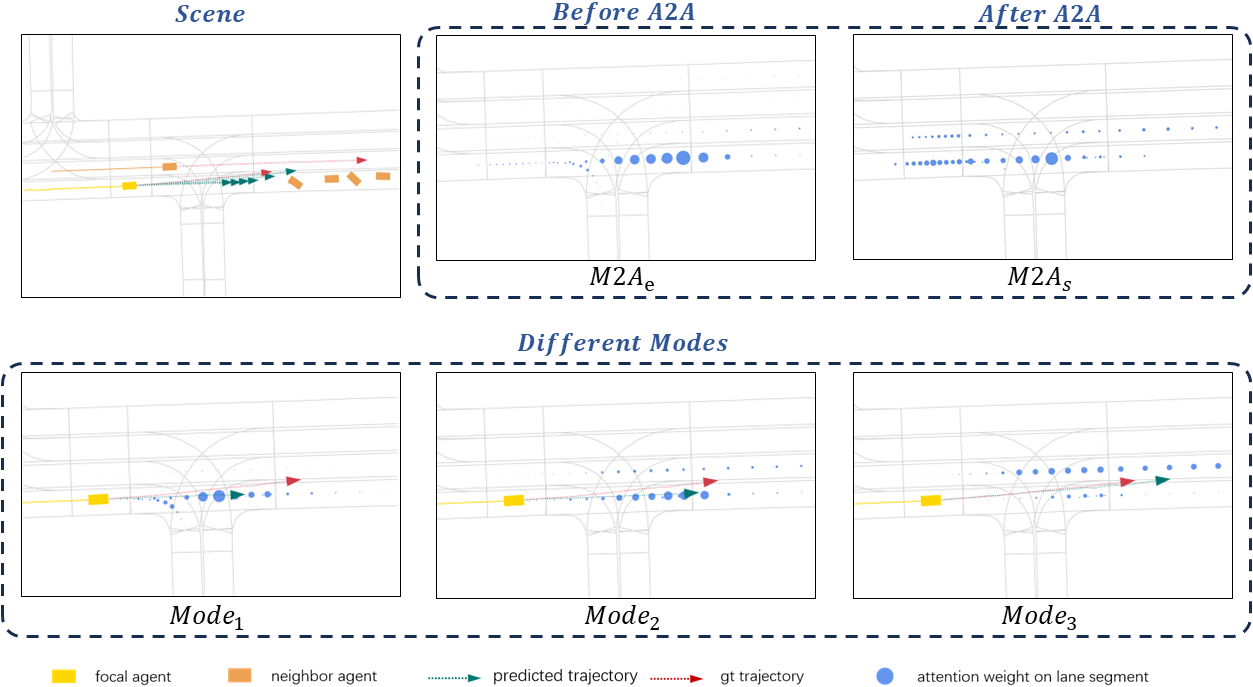}
    \caption{Visualization of the attentions of a focal agent on map nodes at different progressive interaction stages. Larger blue dots indicate higher attention on the map nodes. The last two images in the first row show the changes in attention caused by neighboring trajectories. The images in the second row illustrate the differences in attention for three different modes. The neighbor agents without gt trajectory are static.}
    \label{fig:att_sence}
\end{figure*}


\textcolor{black}{To further validate the rationality of our idea for progressive interactions, we visualize the attentions of a focal agent on map at different stages. As shown in Figure \ref{fig:att_sence}, the first image displays the scene, where the focal agent performs a left lane changing behavior due to the static obstacles from the front and also keeps away from the left driving vehicle. The next two images in the first row illustrate the changes in attention weights on the map nodes before and after the A2A interaction. Note how the weights on the map nodes decrease due to the front agents~(obstacles) after A2A interaction, as well as how the attention successfully captures the potential behavior of left changing lane, as indicated by the newly appeared weights on the left front of the focal agent in the third image. The three images in the bottom demonstrate three predictions and their corresponding attentions on the map in M2A$_{m}$. Clearly, when predicting multiple possible trajectories for the focal agent, M2A$_{m}$ promotes diverse predictions by focusing on different map parts in different predictions.}


\begin{table}
  \caption{Impact of allocation loss and endpoint loss on the AV1 test and validation sets.}
  \centering
  \begin{tabular}{ c c | c c }
    \hline
      allocation  &  endpoint  & test set & val set\\
      loss &  loss & minFDE  & minFDE   \\
    \hline
             &          & 1.198 & 0.986 \\
    $\surd$  &     & 1.200 & 0.961 \\
             & $\surd$  & 1.179 & 0.977 \\
    $\surd$  & $\surd$  & {\bf 1.163} & {\bf 0.959} \\
    \hline
  \end{tabular}
    \label{tab:ablation study of loss}
\end{table}

{\bf Allocation and Endpoint losses.} We further evaluate the effectiveness of the proposed allocation and endpoint losses. 
To balance different loss terms, we set ($\alpha$, $\beta $) in Eq.~(\ref{eq:reg_s2}) to (0, 0.2) or (0.2, 0) or (1, 0.2).
As shown in Table \ref{tab:ablation study of loss}, either endpoint loss or allocation loss is helpful for improving the prediction accuracy. Although on the test set, the performance  of adding allocation loss is not improved, it performs quite well on the validation set, which validates its effectiveness. Table \ref{tab:allocation of 6 modalities} shows the impact of these two losses on the FDE and Hit Rate of each prediction branch. When and only when the endpoint of a trajectory is the nearest one to the endpoint of ground truth, the FDE and Hit Rate will be counted. For convenience, the results are sorted according to the FDE. 
By comparing the first two rows, or the last two rows in Table~\ref{tab:allocation of 6 modalities}, adding the allocation loss significantly reduces the FDE, especially in the 4th and 6th branches. With the help of allocation loss, the branches with fewer training times in the original winner-take-all training strategy can learn from more scenes. 
It is interesting to see that adding the endpoint loss~(E) increases the Hit Rate of the first three main branches, enabling them to predict more scenes. Although the FDE of the first three branches also increases a bit due to more number of scenes are predicted, they are still much smaller than those of other branches, therefore, higher Hit Rate on the first three branches finally leads to the smaller minFDE in the whole dataset.

\begin{table}
  \caption{Comparison of FDE/Hit Rate~(\%) on six prediction branches on AV1 validation set. A, E and “{\bf ---}" means allocation loss, endpoint loss and baseline, respectively. }
  \centering
    \scalebox{0.9}{
  \begin{tabular}{c|c c c c c c}
    \hline
    
    \multirow{2}{*}{loss} & \multicolumn{6}{c}{FDE/HR(\%) on six prediction branches} \\
    \cline{2-7}
                        & 1 & 2 & 3 & 4 & 5 & 6 \\

    \hline

    {\bf ---}   & 0.69/27.4  & 0.82/23.6 & 0.84/22.3 & 1.20/12.0 & 1.54/10.4 & 2.60/4.28 \\
    A      & 0.65/24.2 & 0.72/22.1 & 0.80/20.9 & 1.00/15.1 & 1.57/12.0 & 2.32/5.74 \\
    E           & 0.74/29.4 & 0.84/24.8 & 0.88/22.6 & 1.26/11.0 & 1.49/9.20 & 2.66/2.98   \\
    E+A         & 0.71/27.8 & 0.77/24.0 & 0.88/22.5 & 1.07/12.7 & 1.55/8.90 & 2.54/4.20   \\
    

    \hline
  \end{tabular}
      }

  \label{tab:allocation of 6 modalities}
\end{table}

\begin{figure}[h]
   \centering
   
     \includegraphics[width=3.8cm]{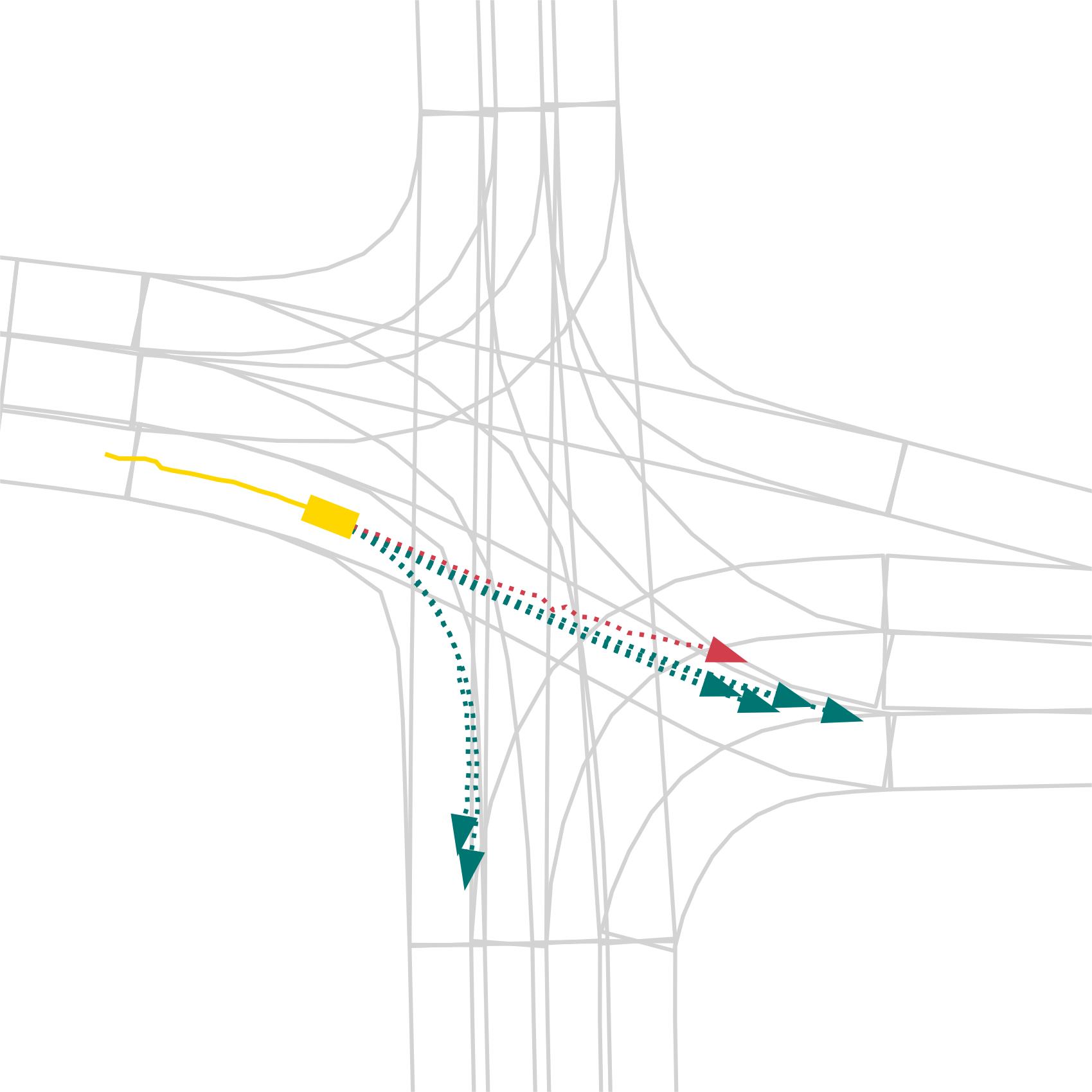}
     \includegraphics[width=3.8cm]{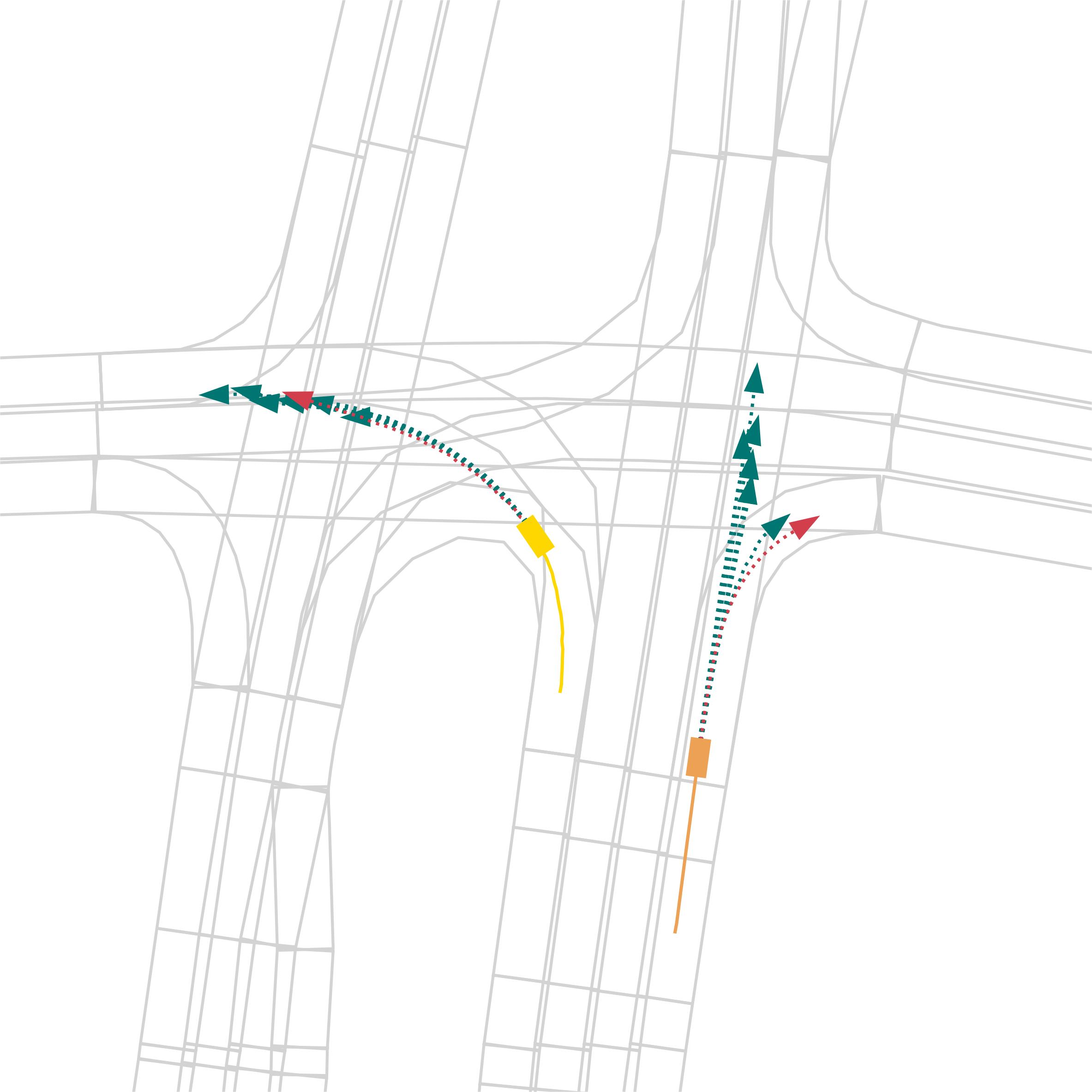}
   \includegraphics[width=3.8cm]{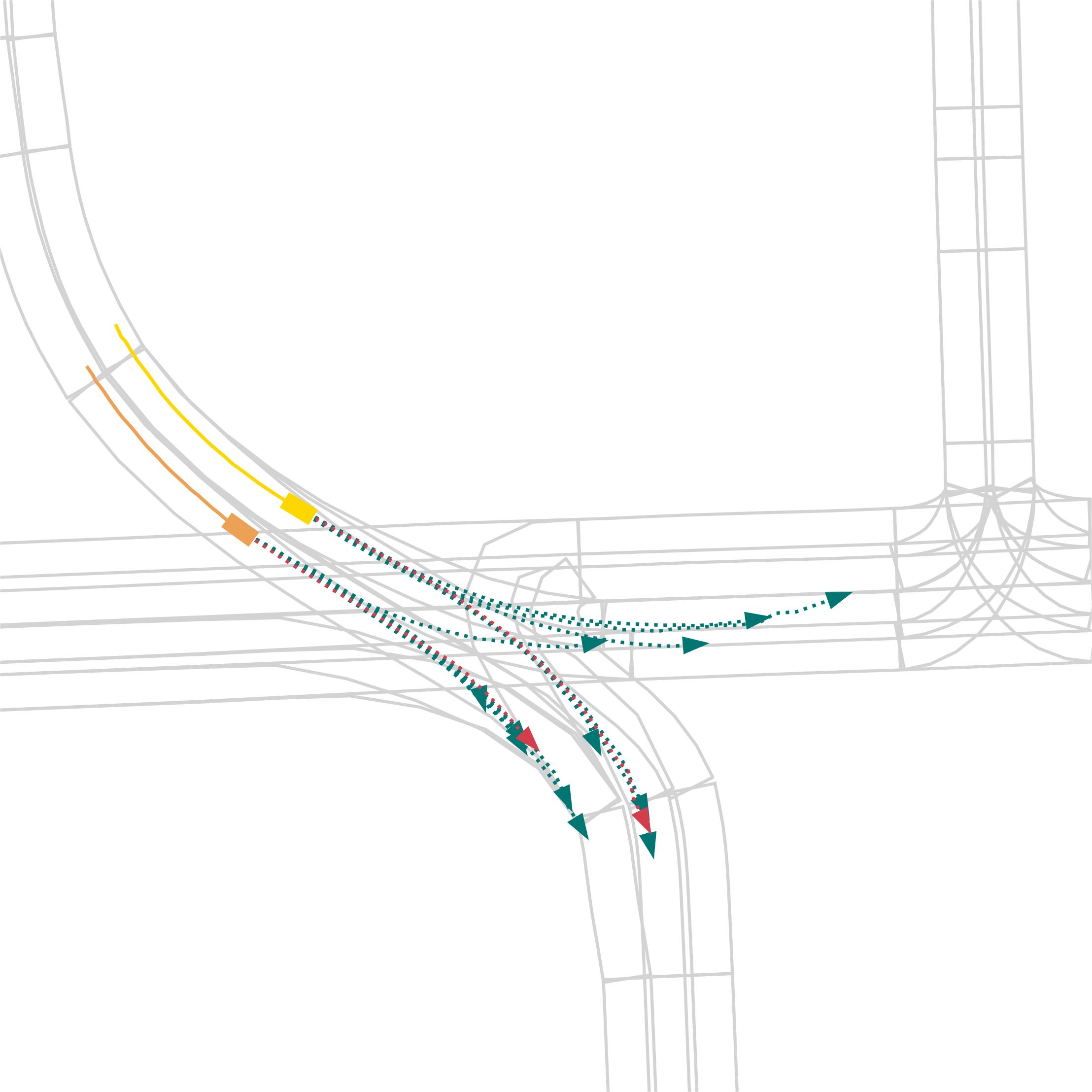}
   \includegraphics[width=3.8cm]{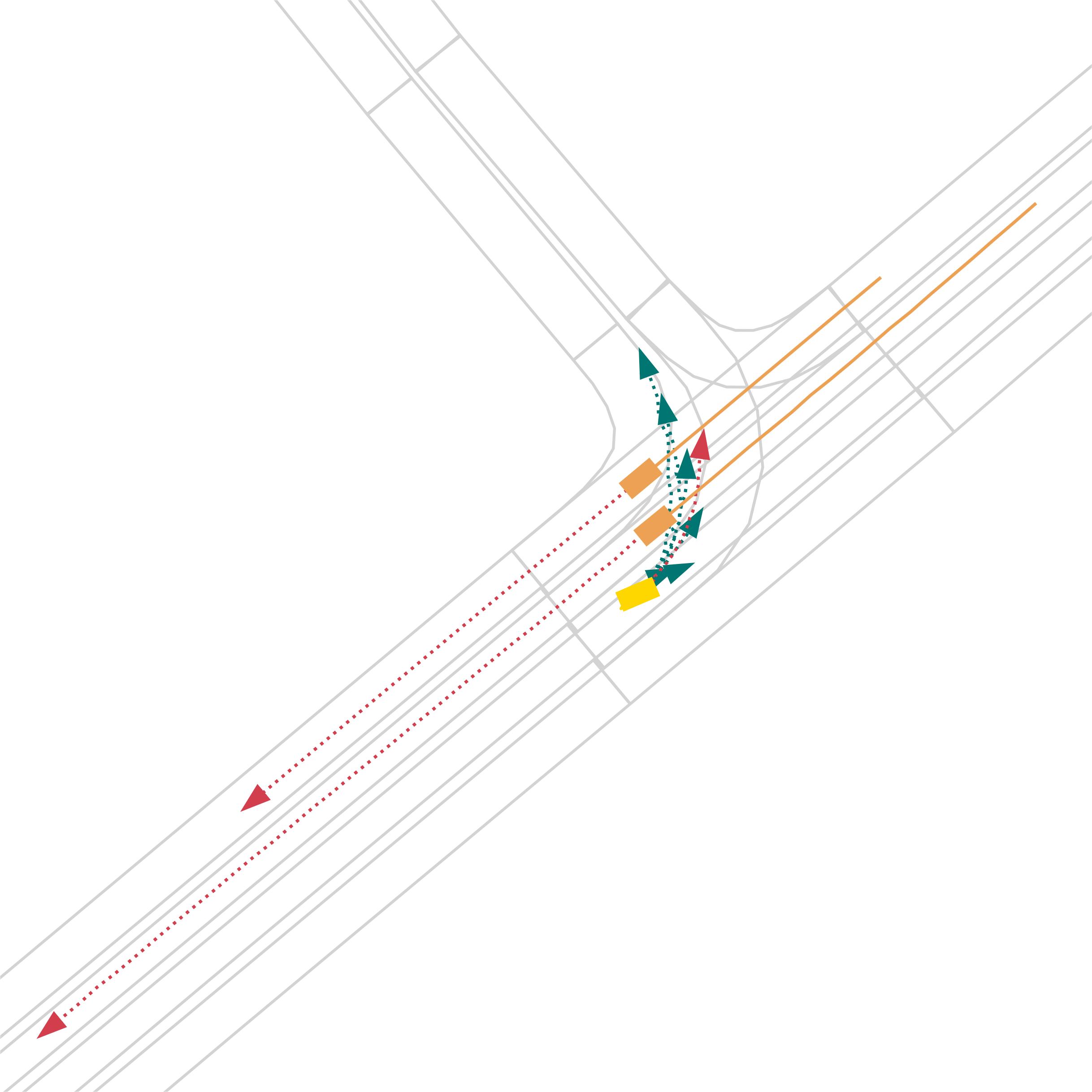}
  
   \caption{Qualitative results of the proposed model on the Argoverse1 validation set. Please refer to Figure~\ref{fig:att_sence} for the meanings of colors and symbols. }
   \label{fig:sence}

 \end{figure}

\subsection{Results}
{\bf Comparison with State-of-the-art.} We compare our method with the state-of-the-art methods in Table~\ref{tab:results on AV1 Leaderboard} and Table~\ref{tab:results on Aargo.org2 Leaderboard}. 
Among the compared methods, LaneGCN~\cite{liang2020learning} is a strong baseline for agent and map interactions by a holistic design of interactions with GCNs, and achieved minFDE of 1.36 and Brier-minFDE of 2.06. The advantages of the proposed progressive interaction over holistic interactions are apparent by comparing our results to LaneGCN. Compared to the existing improvements over LaneGCN with sophisticated interactions, such as DSP~\cite{zhang2022trajectoryDSP}, PAGA~\cite{2022paga}, and GANet~\cite{wang2023ganet}, \textcolor{black}{our method is still competitive, obtaining better results in both minFDE and Brier-minFDE. The proposed ProIn also outperforms other vectorized methods in terms of minFDE and Brier-minFDE metrics, including TPCN\cite{ye2021tpcn} and SceneTransformer~\cite{ngiam2022scene}, as well as the state of the art frameworks for motion forecasting like Wayformer~\cite{nayakanti2022wayformer} and GANet~\cite{wang2023ganet}. Please note that Wayformer uses more feature dimensions and network layers (i.e., higher complexity) to achieve the best minADE than other methods, but is much worse than our method in terms of minFDE. }

We further employ a self-ensemble strategy to improve the prediction accuracy as previous methods did~\cite{MacFormer, varadarajan2022multipath++,HeteroGCN,gilles2022gohome}. We construct eight trajectory decoders as the hydra prediction heads to produce multimodal future trajectories in parallel. Then the k-means algorithm is used to cluster the produced states from these models into 6 categories. The results with self-ensemble are marked by * in Table \ref{tab:results on AV1 Leaderboard} and Table \ref{tab:results on Aargo.org2 Leaderboard}. \textcolor{black}{Among the compared methods, GOHOME~\cite{gilles2022gohome} is specifically designed for MR by sacrificing accuracy. HeteroGCN is based on recurrent GCN, which is biased to long prediction sequence at the cost of higher computation burden. So HeteroGCN achieves the best minADE in AV2 (predicting 5 seconds) while not on the top in AV1 (predicting 3 seconds). Except for these specifically designed methods for some certain metrics, our method beats previous works based on the self-ensemble strategy~(i.e., MacFormer~\cite{MacFormer}, Multipath++~\cite{varadarajan2022multipath++}, HeteroGCN~\cite{HeteroGCN}, GOHOME~\cite{gilles2022gohome}), especially on the minFDE and Brier-minFDE metrics.}

{\bf Qualitative Results.} Qualitative results of our method on the AV1 validation set are presented in Figure \ref{fig:sence}, where only the focal agent and their interacting neighbors are visualized for neatness. As demonstrated, our method can accurately predict multi-modal outcomes in complex traffic scenarios. For instance, the bottom left shows that the focal agent can consider maps at a quite large distance, while in the lower right the agent correctly predicts a turn left action based on neighbor interaction.

\begin{table}
  \caption{Results on the Argoverse 1 Motion Forecasting test set. * denotes using self-ensemble.}
  \centering
  \scalebox{0.82}{
  \begin{tabular}{l | c | c c c c}
  
    \hline
    Method   &  Venue   & b-FDE & minFDE & minADE & MR   \\
    \hline
    LaneGCN\cite{liang2020learning}  
    & ECCV 2020     
    & 2.0584 & 1.3640 & 0.8679 & 16.34  \\
    mmTransformer\cite{liu2021multimodal}   
    &  CVPR 2021    
    & 2.0328 & 1.3383 & 0.8436 & 15.40  \\
    MultiModalTransformer\cite{huang2022multi}   
    &  ICRA 2022    
    & 1.9393 & 1.2905 & 0.8372 & 14.29  \\
    TPCN\cite{ye2021tpcn}
    & CVPR 2021     
    & 1.9286 & 1.2442 & 0.8153 & 13.33  \\
    Scene Transformer\cite{ngiam2022scene}  
    & ICLR 2022     
    & 1.8868 & 1.2321 & 0.8026 & 12.55  \\
    HOME+GOHOME*\cite{gilles2022gohome}
    & ICRA 2022     
    & 1.8601& 1.2919 & 0.8904 &\textbf{8.46} 	\\
    DSP\cite{zhang2022trajectoryDSP}  
    & IROS 2022     
    & 1.8584 & 1.2186 & 0.8194 & 13.03\\
    HiVT-128\cite{zhou2022hivt}  
    & CVPR 2022     
    & 1.8422 & 1.1693 & {0.7735} & 12.67  \\
    Multipath++* \cite{varadarajan2022multipath++}	
    & ICRA 2022      
    & 1.7932 & 1.2144 & 0.7897 & 13.24  \\
    GANet\cite{wang2023ganet}  
    & ICRA 2023 	
    & 1.7899 & 1.1605 & 0.8060 & 11.79\\

    MacFormer* \cite{MacFormer}
    & RAL 2023
    & 1.7667   & 1.2141 & 0.8121 & 12.72  \\
    PAGA\cite{2022paga}  
    & ICRA 2022    
    & 1.7568 & 1.2139 & 0.8014 & {11.43}  \\
    HeteroGCN* \cite{HeteroGCN}	
    & RAL 2023     
    & 1.7512 & 1.1602 & 0.7890 & 11.68  \\
    Wayformer\cite{nayakanti2022wayformer}  
    & ICRA 2023    
    & {1.7408} & {1.1616} & \textbf{{0.7676}} & 11.86  \\
    \hline
    ProIn     &               & {1.7483} &  {1.1554} & 0.8046 & 11.77 \\
    ProIn*     &               & \textbf{1.7076} & \textbf{1.1236} & 0.7762 & 11.64 \\






    \hline
  \end{tabular}}

  \label{tab:results on AV1 Leaderboard}
\end{table}

\begin{table}
  \caption{Results on the Argoverse 2 Motion Forecasting test set. * denotes using self-ensemble.}
  \centering
  \scalebox{1}{
  \begin{tabular}{l | c c c c }
  
    \hline
    Methed      & b-FDE & minFDE & minADE &  MR   \\
    \hline
    THOMAS\cite{gilles2021thomas}       & {2.16} &  {1.51} & {0.88} & {20} \\
    GoRela\cite{cui2022gorela}	    & 2.01 & 1.48 & 0.76 & 22\\
    GANet\cite{wang2023ganet}       & {1.96} &  {1.34} & {0.72} & {17} \\
    BANet*\cite{zhangbanet}           & 1.92 & 1.36 & 0.71 & 19 \\
    MacFormer* \cite{MacFormer}       & 1.90   & 1.38 & 0.70 & 19  \\
    HeteroGCN* \cite{HeteroGCN}       & 1.90 & 1.34 & \textbf{0.69} & 18 \\
    \hline
    ProIn        & {1.93}  & {{1.35}} & {{0.73}}  & {{18}} \\
    ProIn*        & \textbf{1.89}  & \textbf{{1.31}} & {{0.70}}  & \textbf{{17}} \\
    
    \hline
  \end{tabular}}
  \label{tab:results on Aargo.org2 Leaderboard}
\end{table}

\section{Conclusion}

A progressive interaction network is proposed in this paper to better learn agents’ feature representation capturing the scene context information. The proposed progressive interaction network can gradually incorporate map information into agent features at three different stages, and thus can encode map information adapted to social interaction and multi-modal prediction into the agent feature, leading to the improved encoding of scene contextual information. 
Extensive ablations have demonstrated the rationality of the proposed progressive interactions. Experiments have demonstrated significant performance improvement of our method over its counterpart LaneGCN, and our method obtained encouraging results on both Argoverse 1 and Argoverse 2 motion forecasting challenges.

\bibliographystyle{IEEEtran}
\bibliography{egbib}

\begin{thebibliography}{10}
\providecommand{\url}[1]{#1}
\csname url@samestyle\endcsname
\providecommand{\newblock}{\relax}
\providecommand{\bibinfo}[2]{#2}
\providecommand{\BIBentrySTDinterwordspacing}{\spaceskip=0pt\relax}
\providecommand{\BIBentryALTinterwordstretchfactor}{4}
\providecommand{\BIBentryALTinterwordspacing}{\spaceskip=\fontdimen2\font plus
\BIBentryALTinterwordstretchfactor\fontdimen3\font minus
  \fontdimen4\font\relax}
\providecommand{\BIBforeignlanguage}[2]{{%
\expandafter\ifx\csname l@#1\endcsname\relax
\typeout{** WARNING: IEEEtran.bst: No hyphenation pattern has been}%
\typeout{** loaded for the language `#1'. Using the pattern for}%
\typeout{** the default language instead.}%
\else
\language=\csname l@#1\endcsname
\fi
#2}}
\providecommand{\BIBdecl}{\relax}
\BIBdecl

\bibitem{trajgan}
X.~Li, G.~Rosman, I.~Gilitschenski, C.-I. Vasile, J.~A. DeCastro, S.~Karaman,
  and D.~Rus, ``Vehicle trajectory prediction using generative adversarial
  network with temporal logic syntax tree features,'' \emph{IEEE Robotics and
  Automation Letters}, vol.~6, no.~2, pp. 3459--3466, 2021.

\bibitem{aandm2a}
L.~Wang, T.~Wu, H.~Fu, L.~Xiao, Z.~Wang, and B.~Dai, ``Multiple contextual cues
  integrated trajectory prediction for autonomous driving,'' \emph{IEEE
  Robotics and Automation Letters}, vol.~6, no.~4, pp. 6844--6851, 2021.

\bibitem{casas2018intentnet}
S.~Casas, W.~Luo, and R.~Urtasun, ``Intentnet: Learning to predict intention
  from raw sensor data,'' in \emph{Conference on Robot Learning}.\hskip 1em
  plus 0.5em minus 0.4em\relax PMLR, 2018, pp. 947--956.

\bibitem{hong2019rules}
J.~Hong, B.~Sapp, and J.~Philbin, ``Rules of the road: Predicting driving
  behavior with a convolutional model of semantic interactions,'' in
  \emph{Proceedings of the IEEE/CVF Conference on Computer Vision and Pattern
  Recognition}, 2019, pp. 8454--8462.

\bibitem{gilles2021home}
T.~Gilles, S.~Sabatini, D.~Tsishkou, B.~Stanciulescu, and F.~Moutarde, ``Home:
  Heatmap output for future motion estimation,'' in \emph{2021 IEEE
  International Intelligent Transportation Systems Conference (ITSC)}.\hskip
  1em plus 0.5em minus 0.4em\relax IEEE, 2021, pp. 500--507.

\bibitem{zhang2022trajectoryDSP}
L.~Zhang, P.~Li, J.~Chen, and S.~Shen, ``Trajectory prediction with graph-based
  dual-scale context fusion,'' in \emph{2022 IEEE/RSJ International Conference
  on Intelligent Robots and Systems (IROS)}.\hskip 1em plus 0.5em minus
  0.4em\relax IEEE, 2022, pp. 11\,374--11\,381.

\bibitem{gao2020vectornet}
J.~Gao, C.~Sun, H.~Zhao, Y.~Shen, D.~Anguelov, C.~Li, and C.~Schmid,
  ``Vectornet: Encoding hd maps and agent dynamics from vectorized
  representation,'' in \emph{Proceedings of the IEEE/CVF Conference on Computer
  Vision and Pattern Recognition}, 2020, pp. 11\,525--11\,533.

\bibitem{TPA_laneloss}
S.~Kim, H.~Jeon, J.~W. Choi, and D.~Kum, ``Diverse multiple trajectory
  prediction using a two-stage prediction network trained with lane loss,''
  \emph{IEEE Robotics and Automation Letters}, vol.~8, no.~4, pp. 2038--2045,
  2023.

\bibitem{liang2020learning}
M.~Liang, B.~Yang, R.~Hu, Y.~Chen, R.~Liao, S.~Feng, and R.~Urtasun, ``Learning
  lane graph representations for motion forecasting,'' in \emph{European
  Conference on Computer Vision}.\hskip 1em plus 0.5em minus 0.4em\relax
  Springer, 2020, pp. 541--556.

\bibitem{2022paga}
F.~Da and Y.~Zhang, ``Path-aware graph attention for hd maps in motion
  prediction,'' in \emph{2022 International Conference on Robotics and
  Automation (ICRA)}.\hskip 1em plus 0.5em minus 0.4em\relax IEEE, 2022, pp.
  6430--6436.

\bibitem{wang2023ganet}
M.~Wang, X.~Zhu, C.~Yu, W.~Li, Y.~Ma, R.~Jin, X.~Ren, D.~Ren, M.~Wang, and
  W.~Yang, ``Ganet: Goal area network for motion forecasting,'' in \emph{2023
  IEEE International Conference on Robotics and Automation (ICRA)}, 2023, pp.
  1609--1615.

\bibitem{gu2021densetnt}
J.~Gu, C.~Sun, and H.~Zhao, ``Densetnt: End-to-end trajectory prediction from
  dense goal sets,'' in \emph{Proceedings of the IEEE/CVF International
  Conference on Computer Vision}, 2021, pp. 15\,303--15\,312.

\bibitem{gilles2022gohome}
T.~Gilles, S.~Sabatini, D.~Tsishkou, B.~Stanciulescu, and F.~Moutarde,
  ``Gohome: Graph-oriented heatmap output for future motion estimation,'' in
  \emph{2022 International Conference on Robotics and Automation (ICRA)}.\hskip
  1em plus 0.5em minus 0.4em\relax IEEE, 2022, pp. 9107--9114.

\bibitem{HeteroGCN}
X.~Gao, X.~Jia, Y.~Li, and H.~Xiong, ``Dynamic scenario representation learning
  for motion forecasting with heterogeneous graph convolutional recurrent
  networks,'' \emph{IEEE Robotics and Automation Letters}, vol.~8, no.~5, pp.
  2946--2953, 2023.

\bibitem{liu2021multimodal}
Y.~Liu, J.~Zhang, L.~Fang, Q.~Jiang, and B.~Zhou, ``Multimodal motion
  prediction with stacked transformers,'' in \emph{Proceedings of the IEEE/CVF
  Conference on Computer Vision and Pattern Recognition}, 2021, pp. 7577--7586.

\bibitem{ye2021tpcn}
M.~Ye, T.~Cao, and Q.~Chen, ``Tpcn: Temporal point cloud networks for motion
  forecasting,'' in \emph{Proceedings of the IEEE/CVF Conference on Computer
  Vision and Pattern Recognition}, 2021, pp. 11\,318--11\,327.

\bibitem{ye2023dcms}
M.~Ye, J.~Xu, X.~Xu, T.~Wang, T.~Cao, and Q.~Chen, ``Dcms: Motion forecasting
  with dual consistency and multi-pseudo-target supervision,'' 2023.

\bibitem{vaswani2017attention}
A.~Vaswani, N.~Shazeer, N.~Parmar, J.~Uszkoreit, L.~Jones, A.~N. Gomez,
  {\L}.~Kaiser, and I.~Polosukhin, ``Attention is all you need,''
  \emph{Advances in neural information processing systems}, vol.~30, 2017.

\bibitem{Xiaoyu}
X.~Mo, Y.~Xing, H.~Liu, and C.~Lv, ``Map-adaptive multimodal trajectory
  prediction using hierarchical graph neural networks,'' \emph{IEEE Robotics
  and Automation Letters}, vol.~8, no.~6, pp. 3685--3692, 2023.

\bibitem{laneheadloss}
R.~Greer, N.~Deo, and M.~Trivedi, ``Trajectory prediction in autonomous driving
  with a lane heading auxiliary loss,'' \emph{IEEE Robotics and Automation
  Letters}, vol.~6, no.~3, pp. 4907--4914, 2021.

\bibitem{nayakanti2022wayformer}
N.~Nayakanti, R.~Al-Rfou, A.~Zhou, K.~Goel, K.~S. Refaat, and B.~Sapp,
  ``Wayformer: Motion forecasting via simple \& efficient attention networks,''
  in \emph{2023 IEEE International Conference on Robotics and Automation
  (ICRA)}, 2023, pp. 2980--2987.

\bibitem{huang2022multi}
Z.~Huang, X.~Mo, and C.~Lv, ``Multi-modal motion prediction with
  transformer-based neural network for autonomous driving,'' in \emph{2022
  International Conference on Robotics and Automation (ICRA)}.\hskip 1em plus
  0.5em minus 0.4em\relax IEEE, 2022, pp. 2605--2611.

\bibitem{makansi2019overcoming}
O.~Makansi, E.~Ilg, O.~Cicek, and T.~Brox, ``Overcoming limitations of mixture
  density networks: A sampling and fitting framework for multimodal future
  prediction,'' in \emph{Proceedings of the IEEE/CVF Conference on Computer
  Vision and Pattern Recognition}, 2019, pp. 7144--7153.

\bibitem{zhou2022hivt}
Z.~Zhou, L.~Ye, J.~Wang, K.~Wu, and K.~Lu, ``Hivt: Hierarchical vector
  transformer for multi-agent motion prediction,'' in \emph{Proceedings of the
  IEEE/CVF Conference on Computer Vision and Pattern Recognition}, 2022, pp.
  8823--8833.

\bibitem{ngiam2022scene}
J.~Ngiam, V.~Vasudevan, B.~Caine, Z.~Zhang, H.-T.~L. Chiang, J.~Ling,
  R.~Roelofs, A.~Bewley, C.~Liu, A.~Venugopal \emph{et~al.}, ``Scene
  transformer: A unified architecture for predicting future trajectories of
  multiple agents,'' in \emph{International Conference on Learning
  Representations}.

\bibitem{MacFormer}
C.~Feng, H.~Zhou, H.~Lin, Z.~Zhang, Z.~Xu, C.~Zhang, B.~Zhou, and S.~Shen,
  ``Macformer: Map-agent coupled transformer for real-time and robust
  trajectory prediction,'' \emph{IEEE Robotics and Automation Letters}, vol.~8,
  no.~10, pp. 6795--6802, 2023.

\bibitem{Wang_2023_CVPR}
X.~Wang, T.~Su, F.~Da, and X.~Yang, ``Prophnet: Efficient agent-centric motion
  forecasting with anchor-informed proposals,'' in \emph{Proceedings of the
  IEEE/CVF Conference on Computer Vision and Pattern Recognition (CVPR)}, 2023,
  pp. 21\,995--22\,003.

\bibitem{Zhou_2023_CVPR}
Z.~Zhou, J.~Wang, Y.-H. Li, and Y.-K. Huang, ``Query-centric trajectory
  prediction,'' in \emph{Proceedings of the IEEE/CVF Conference on Computer
  Vision and Pattern Recognition (CVPR)}, 2023, pp. 17\,863--17\,873.

\bibitem{gilles2021thomas}
T.~Gilles, S.~Sabatini, D.~Tsishkou, B.~Stanciulescu, and F.~Moutarde,
  ``Thomas: Trajectory heatmap output with learned multi-agent sampling,'' in
  \emph{International Conference on Learning Representations}, 2021.

\bibitem{chang2019argoverse}
M.-F. Chang, J.~Lambert, P.~Sangkloy, J.~Singh, S.~Bak, A.~Hartnett, D.~Wang,
  P.~Carr, S.~Lucey, D.~Ramanan \emph{et~al.}, ``Argoverse: 3d tracking and
  forecasting with rich maps,'' in \emph{Proceedings of the IEEE/CVF Conference
  on Computer Vision and Pattern Recognition}, 2019, pp. 8748--8757.

\bibitem{Argoverse2}
B.~Wilson, W.~Qi, T.~Agarwal, J.~Lambert, J.~Singh, S.~Khandelwal, B.~Pan,
  R.~Kumar, A.~Hartnett, J.~K. Pontes, D.~Ramanan, P.~Carr, and J.~Hays,
  ``Argoverse 2: Next generation datasets for self-driving perception and
  forecasting,'' in \emph{Proceedings of the Neural Information Processing
  Systems Track on Datasets and Benchmarks (NeurIPS Datasets and Benchmarks
  2021)}, 2021.

\bibitem{argo.org}
A.~M.~F. Competition,
  \url{https://eval.ai/web/challenges/challenge-page/454/leaderboard/1279}
  Accessed 2022-11-11.

\bibitem{argo2.org}
A.~. M.~F. Competition,
  \url{https://eval.ai/web/challenges/challenge-page/1719/leaderboard/4098}
  Accessed 2022-11-11.

\bibitem{kingma2014adam}
D.~P. Kingma and J.~Ba, ``Adam: A method for stochastic optimization,''
  \emph{arXiv preprint arXiv:1412.6980}, 2014.

\bibitem{varadarajan2022multipath++}
B.~Varadarajan, A.~Hefny, A.~Srivastava, K.~S. Refaat, N.~Nayakanti,
  A.~Cornman, K.~Chen, B.~Douillard, C.~P. Lam, D.~Anguelov \emph{et~al.},
  ``Multipath++: Efficient information fusion and trajectory aggregation for
  behavior prediction,'' in \emph{2022 International Conference on Robotics and
  Automation (ICRA)}.\hskip 1em plus 0.5em minus 0.4em\relax IEEE, 2022, pp.
  7814--7821.

\bibitem{cui2022gorela}
A.~Cui, S.~Casas, K.~Wong, S.~Suo, and R.~Urtasun, ``Gorela: Go relative for
  viewpoint-invariant motion forecasting,'' \emph{arXiv preprint
  arXiv:2211.02545}, 2022.

\bibitem{zhangbanet}
C.~Zhang, H.~Sun, C.~Chen, and Y.~Guo, ``Banet: Motion forecasting with
  boundary aware network,'' \emph{arXiv preprint arXiv:2206.07934}, 2022.

\end{thebibliography}


\end{document}